\documentclass[fleqn,10pt]{wlscirep}
\usepackage[utf8]{inputenc}
\usepackage[T1]{fontenc}
\usepackage{placeins}
\title{Improving Quality Control of Whole Slide Images by Explicit Artifact Augmentation}

\author[1,2*]{Artur Jurgas}
\author[1,2]{Marek Wodzinski}
\author[3]{Marina D'Amato}
\author[3]{Jeroen van der Laak}
\author[2,4]{Manfredo Atzori}
\author[2,5]{Henning M\"{u}ller}
\affil[1]{AGH University of Krakow, Faculty of Electrical Engineering, Automatics, Computer Science and Biomedical Engineering, 30059 Krakow, Poland}
\affil[2]{University of Applied Sciences Western Switzerland (HES-SO), Institute of Informatics, 3960 Sierre, Switzerland}
\affil[3]{Radboud University Medical Center, Nijmegen, The Netherlands}
\affil[4]{Department of Neuroscience, University of Padova, Padova, Italy}
\affil[5]{Medical Faculty, University of Geneva, Geneva, Switzerland}

\affil[*]{arjurgas@agh.edu.pl}


\begin{abstract}
The problem of artifacts in whole slide image acquisition, prevalent in both clinical workflows and research-oriented settings, necessitates human intervention and re-scanning. Overcoming this challenge requires developing quality control algorithms, that are hindered by the limited availability of relevant annotated data in histopathology. The manual annotation of ground-truth for artifact detection methods is expensive and time-consuming. This work addresses the issue by proposing a method dedicated to augmenting whole slide images with artifacts. The tool seamlessly generates and blends artifacts from an external library to a given histopathology dataset. The augmented datasets are then utilized to train artifact classification methods. The evaluation shows their usefulness in classification of the artifacts, where they show an improvement from 0.10 to 0.01 AUROC depending on the artifact type. The framework, model, weights, and ground-truth annotations are freely released to facilitate open science and reproducible research.
\end{abstract}
\begin{document}

\flushbottom
\maketitle

\thispagestyle{empty}

\section*{Introduction}

\begin{figure}[h]
    \centering
    \includegraphics[width=1\linewidth]{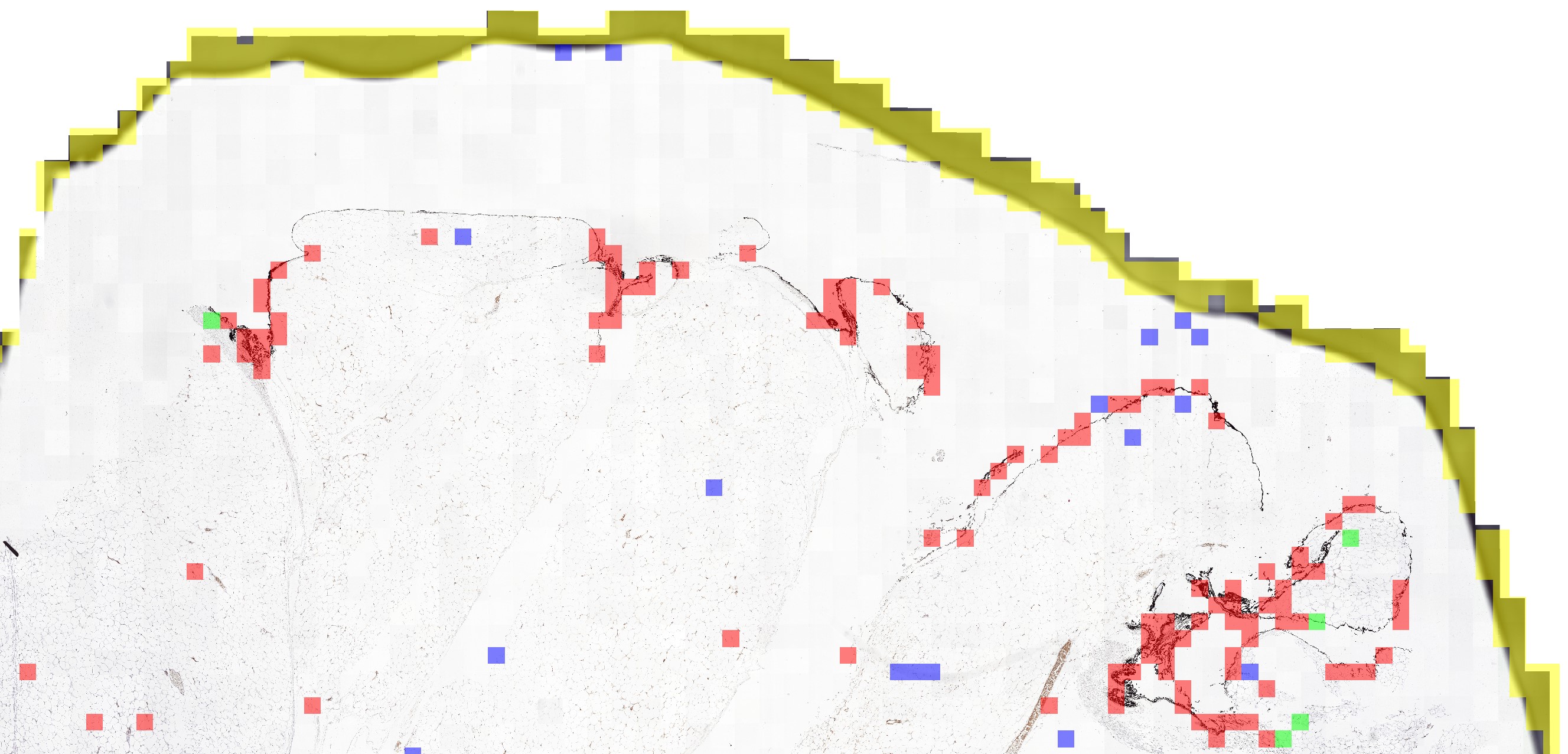}
    \caption{Aggregation of a whole slide image patch-level classification. Color meaning: (yellow) marker, (red) ink, (blue) dust, (green) tissue folding.}
    \label{fig:example-inference}
\end{figure}

With advancements in medical imaging techniques, histopathology remains an irreplaceable cornerstone, offering vital information for disease diagnosis, treatment decisions, and prognosis evaluation. This discipline forms the foundations of various medical specialties, including pathology, oncology, dermatology, and more. With the continued rise of computed histopathology, it carries an increased importance of quality control (QC) systems \cite{Brixtel2022WholeSI}.

Histopathology not only offers a microscopic view of tissue structure but thanks to dyes that stain different structures, can also impart invaluable information about the tissue's composition. This unique combination of ultra-high resolution and color detail distinguishes histopathology from other imaging modalities. However, due to its intricate nature, histopathology workflows are susceptible to various artifacts, like shown in Fig~\ref{fig:example-inference}. The artifacts are unwanted structures or inconsistencies that obstruct the clear visualization of tissue structures \cite{khan_artifacts_2014}.

Artifacts in histopathological images come from various sources, such as flaws in slide preparation, imperfect aperture settings during imaging. They can also come from improperly carrying out staining procedures, especially when it comes to immunohistochemistry dyes \cite{elias_special_1989,tsutsumi_pitfalls_2021}. Inconsistent staining, caused by variations in reagent concentrations, incubation times, or temperature fluctuations, can result in uneven coloration, obscuring critical cellular details \cite{taqi_review_2018, ekundina_common_2015}. Even the most subtle discrepancies in these processes can give rise to artifacts within Whole Slide Images (WSIs), which can potentially compromise the accuracy of medical diagnoses.

Given the significance of histopathology in contemporary medical workflows, it is imperative to ensure that the acquired images are of the highest quality possible. The need for rigorous quality control (QC) mechanisms in histopathology is evident, as the precision of medical diagnoses, treatment planning, and patient outcomes hinge on the reliability of these images. This includes early detection, localization, and classification of unwanted structures on the image, especially if they obstruct the view of the tissue being examined. Structures like that need to be recognized and acknowledged, ideally during the acquisition process as early as possible. The problem is further complicated by the size of the image, as well as the high heterogeneity of the artifacts themselves \cite{kanwal_devil_2022}. This paper focuses on the development and application of a fully end-to-end pipeline for training deep learning-based quality control systems that addresses the challenge of artifact detection in histopathological WSIs, enhancing the reliability and accuracy of medical interpretations.

Recent developments in histopathology QC systems have shown promising strides towards automated assessment. Notably, the semi-automatic HistoQC software \cite{janowczyk_histoqc_2019, chen_assessment_2021} facilitates the automatic evaluation of WSIs for quality assessment (e.g., pen markings, air bubbles, blur), yet several challenges still persist. A primary concern arises from the high heterogeneity observed in clinical datasets, necessitating advanced artifact detection algorithms for each artifact type. This leads to significant time demands in both development and inference phases. 

\begin{figure}[h]
\centering%
\begin{minipage}{0.45\textwidth}
  \includegraphics[width=\linewidth]{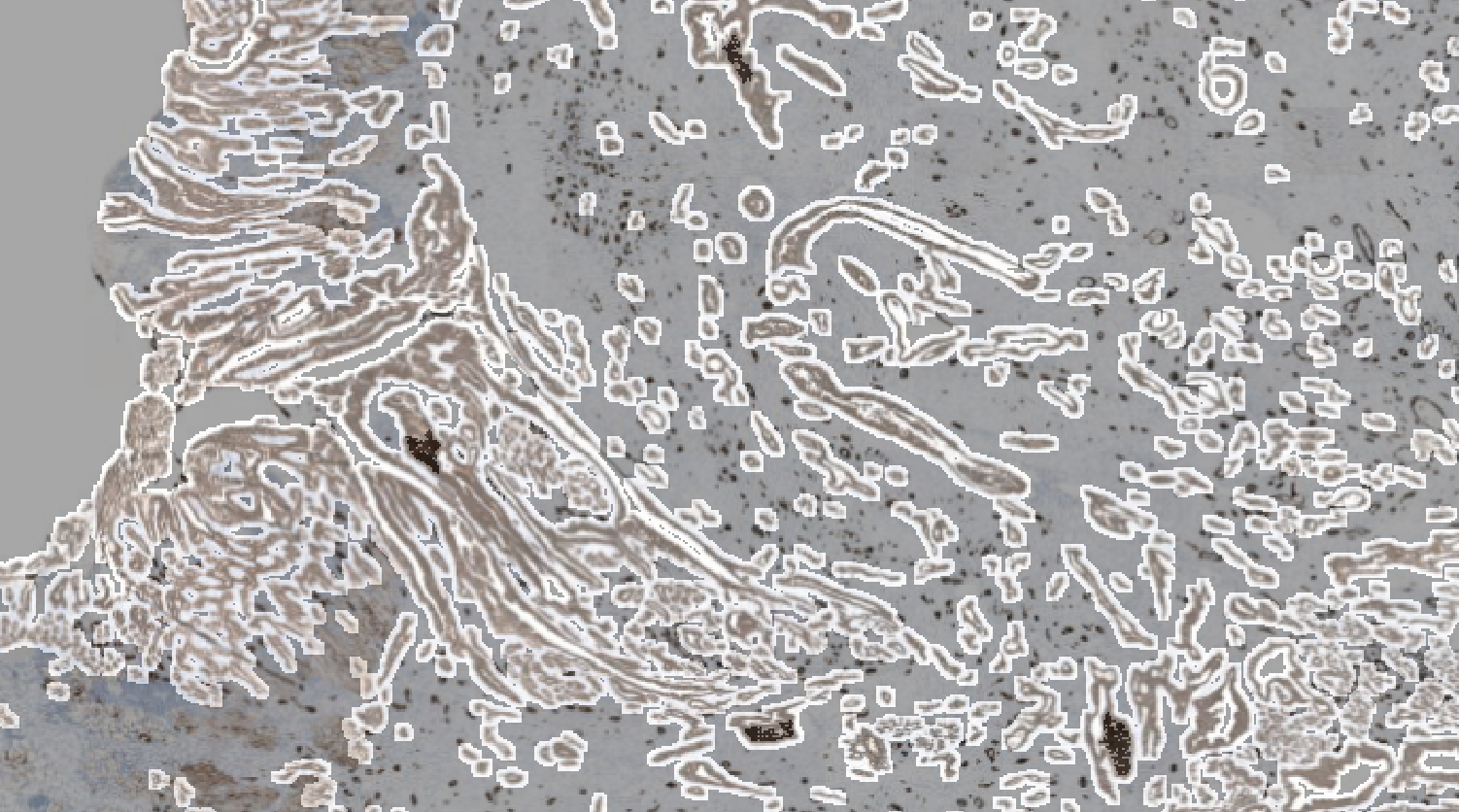}
\end{minipage}%
\hspace{0.02\textwidth}%
\begin{minipage}{0.45\textwidth}
  \includegraphics[width=\linewidth]{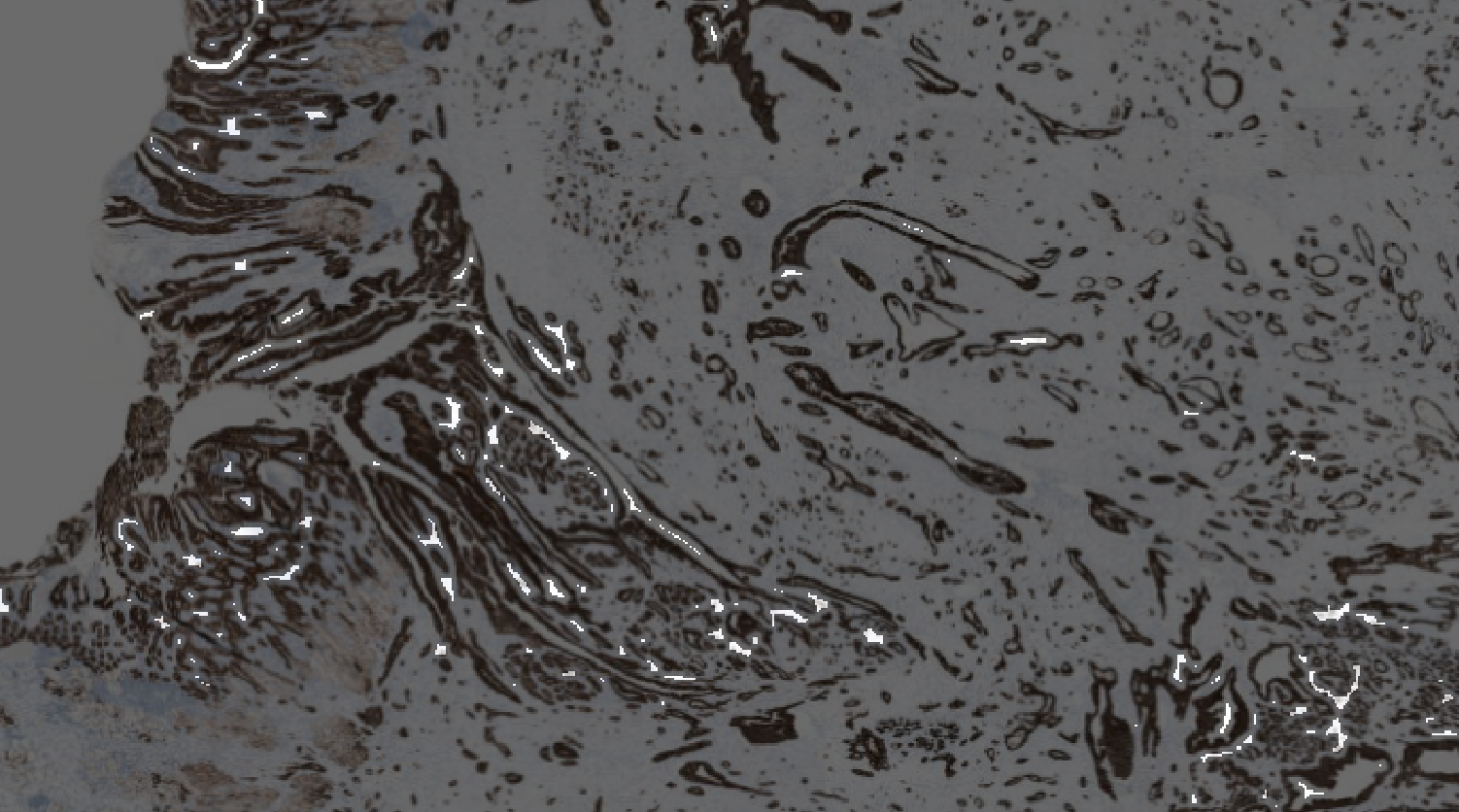}
\end{minipage}%
\caption{Example of failure of a quality control algorithm in identifying artifacts within an IHC image: (left) coverslip edge, (right) small objects (e.g., dust).}
\label{fig:histoqc}
\end{figure}

Most algorithms' effectiveness is prominent in the context of Hematoxylin and Eosin (H\&E) staining, but can exhibit limitations when applied to Immunohistochemistry (IHC) staining. This is due to those dyes' unique challenges and variations, as presented in Figure~\ref{fig:histoqc}. Moreover, most software is parametrized, requires examples of artifacts, and training classification models for each inference. As per the official documentation of HistoQC \cite{noauthor_histoqc_nodate}, it is also recommended to work with images at 1.25x magnification; otherwise the processing can become infeasible. HistoQC is a great achievement for an end-to-end WSI analysis platform, where our method could potentially improve the artifact detection part of the pipeline.

Other authors have explored learning-based QC systems \cite{Campanella2017TowardsML, Senaras2018DeepFocusDO}, but their capabilities are largely confined to detecting blur and out-of-focus artifacts, which are relatively easier to generate synthetically where they convolve parts of an image with a Gaussian function. In a supervised, learning-based approach \cite{smit_quality_2021}, a proprietary labeled dataset was employed for training, albeit with considerable cost and limited generalizability owing to extensive manual annotations. A similar work by \cite{foucart_artifact_2018} leveraged deep residual networks but relied on a modest manually annotated dataset. The critical role of QC in learning-based histopathology methods is underscored in \cite{SchmigMarkiefka2021QualityCS, Wang2021StressTP}. Those works include the generation of artifacts for stress-testing already trained deep learning models for evaluation. Synthetic artifacts are a prevalent focus there, with fewer non-synthetic artifacts addressed.

\subsection*{Contribution}
This work presents a novel data augmentation method that extends high-resolution datasets with seamlessly blended real artifacts, fostering more realistic histopathological analysis challenges. We propose a methodology that involves the extraction of a representative sample of annotated artifacts, utilizing them to generate synthetic, realistic datasets for artifact detection and classification. By reducing the reliance on extensive professional annotations, this approach minimizes costs and enhances results. It also enables easy fine-tuning to accommodate specific institutional settings, allowing researchers to personalize and optimize artifact detection models. Additionally, the study demonstrates how this data generation method enhances the generalization of automatic, learning-based QC methods, resulting in improved performance and robustness.

\section*{Results}

\subsection*{Datasets}
The datasets employed in this study offer a diverse and comprehensive representation of histopathological artifacts. The ACROBAT challenge dataset \cite{Weitz2022ACROBATA} comprises digitalized Whole Slide Images (WSIs) from FFPE surgical resection specimens of female primary breast cancer patients. Captured at 40X magnification (0.23 $\mu$m per pixel), these images, obtained using Hamamatsu Nanozoomer XR or Nanozoomer S360 scanners, exhibit a rich variety of artifacts. The evaluation focused on the validation subset, consisting of 100 cases equally divided between H\&E and IHC-stained images.

Similarly, the ANHIR challenge dataset \cite{Borovec2020ANHIRAN} provides a wide-ranging collection encompassing various tissues and pathological conditions, including lesions, lung lobes, mammary glands, colon adenocarcinoma (COAD), mice kidney tissue, gastric mucosa, gastric adenocarcinoma tissue, breast tissue, and kidney tissue. This dataset incorporates diverse staining techniques, employing stains such as Clara cell 10 protein, proSPC, H\&E, Ki-67, PECAM-1, HER-2/neu, ER, PR, cytokeratin, and podocin. Acquired from various microscopy setups and scanners like Zeiss, Leica, 3DHistec, and NanoZoomer, the ANHIR dataset exhibits high heterogeneity, encompassing magnifications ranging from 10x to 40x and pixel sizes from 0.174 $\mu$m/pixel to 2.294 $\mu$m/pixel.

\begin{figure}[!h]
    \centering
    \includegraphics[width=0.9\linewidth]{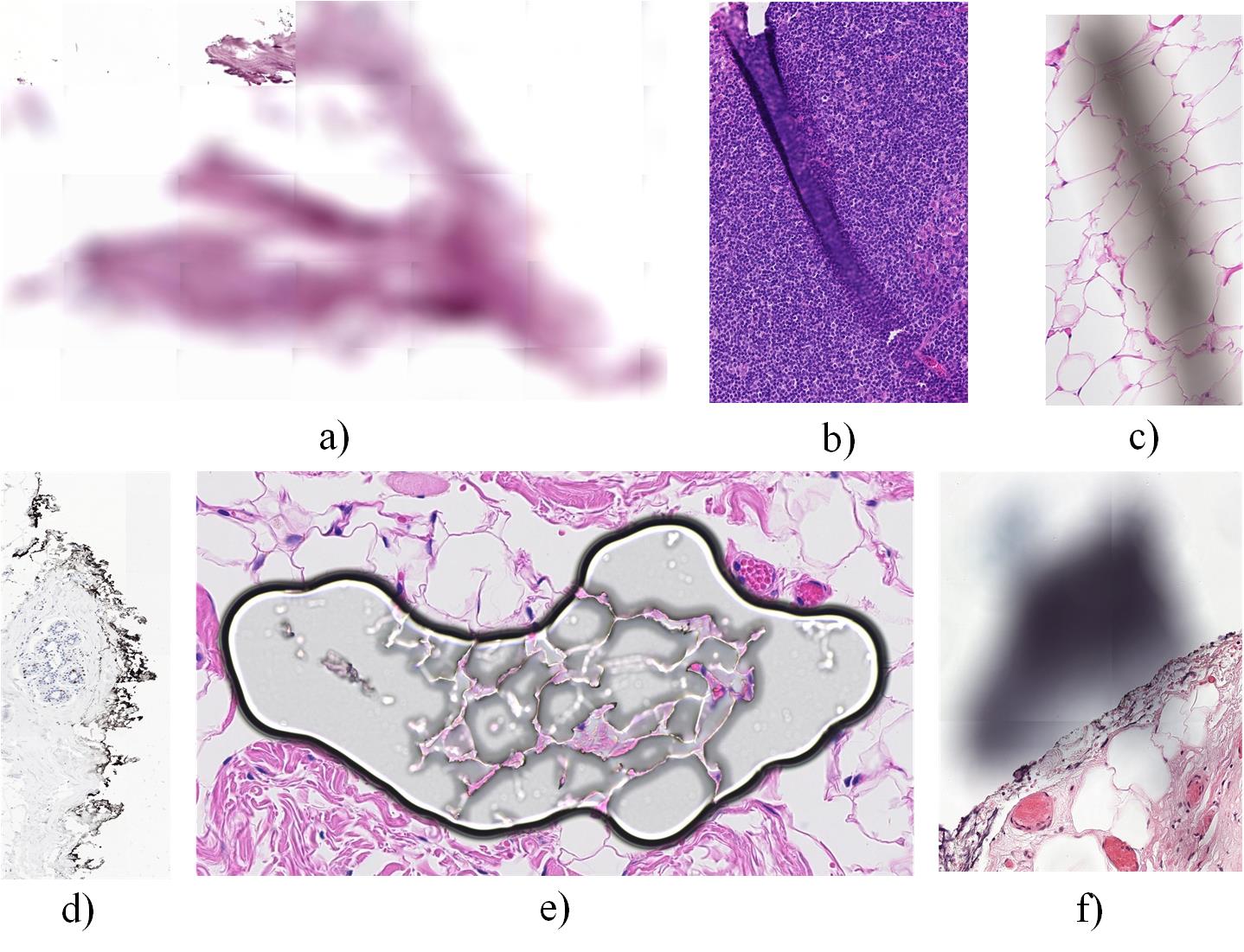}
    \caption{Examples of artifacts from the considered datasets:  a) focus, b) tissue, c) dust, d) ink, e) air, f) marker.}
    \label{fig:rb-examples}
\end{figure}

Additionally, the Radboud University dataset, provided for evaluation purposes, stands as the largest both in artifact count and resolution. This dataset features professionally annotated artifacts in WSIs stained with both H\&E and IHC dyes. Exemplary artifacts are present in Figure~\ref{fig:rb-examples}. Spanning various tissue types, including bone marrow, breast tissue, colon tissue, pancreas tissue, diffuse large B-cell lymphoma (DLBCL), and images from the CAMELYON dataset \cite{litjens_1399_2018}, each tissue type is characterized by different staining types, contributing to the dataset's richness and complexity. The Radboud University dataset offers a substantial resource for evaluating and validating the proposed quality control and segmentation methodologies in diverse histopathological contexts.

Selected artifacts are as follows: 
(i) Air. Not strictly connected to the tissue. Due to the
fact that only a portion of the air bubble is frequently visible in the image, it assumes an open shape, often deviating from a complete circle. 
(ii) Dust. Small particles or debris that can inadvertently appear on the slides during the preparation or scanning process. It appears both on the foreground and background of the WSI.
(iii) Tissue. Folded or creased tissue sections that can result from various factors such as handling, processing, or mounting of the tissue slides.
(iv) Ink. Irregularities in the distribution or application of ink or staining agents on tissue slides.
(v) Marker. Annotations, such as crosses or other symbols, typically located near the corners or edges of the slide.
(vi) Focus. This artifact occurs when the focal plane of the microscope is not precisely aligned with the tissue section being captured, resulting in blurred or out-of-focus areas.
We summarize acquired data in Tab~\ref{tab:data_summary}. 

\begin{table}
    \centering
    \caption{Datasets used in the study with their respective characteristics.}
    \begin{tabular}{ccccccccc}
        \hline
         & & \multicolumn{7}{c}{Number of artifacts}\\
         Dataset&  Source& Air&  Dust&  Tissue&  Ink&  Marker& Focus& All\\
         \hline
         ACROBAT& Self& 67&  137&  89&  87&  97& 14& 491\\
         ANHIR& Self& 17&  130&  62&  44&  19& 25& 297\\
         Radboud& Professional& 149&  398&  1469&  456&  50& 96& 2618\\
         \hline
    \end{tabular}
    \label{tab:data_summary}
\end{table}

\subsection*{Experimental Setup}
The experimental setup utilized Nvidia Tesla A100 graphics cards with 400W TDP and 40 GB of memory on the PLGrid HPC cluster Athena for model training. In our experiments, we employed deep learning models trained on different datasets, denoted by shorthand notations. Models trained exclusively on annotated data from the ACROBAT dataset are referenced as $\mathbf{ACR}$, while models trained on an augmented version of the ACROBAT dataset are denoted as $\mathbf{ACR'}$. Similarly, with $\mathbf{ANH}$, $\mathbf{ANH'}$ for ANHIR and $\mathbf{RB}$, $\mathbf{RB'}$ for Radboud. Additionally, we evaluated models trained on ACROBAT datasets on the ANHIR dataset's annotations to analyze generalizability. Those models are denoted as $\mathbf{ACR_{anh}}$ and, $\mathbf{ACR'_{anh}}$ respectively. When training the models on a dataset from the Radboud University, we present two approaches: (i) while having the full model set to trainable - $\mathbf{RB}$, and (ii) only the last layers unfrozen - $\mathbf{RB_s}$.

\subsection*{Classification}
\begin{figure}[!h]
\centering%
\begin{minipage}{0.45\textwidth}
  \includegraphics[width=\linewidth]{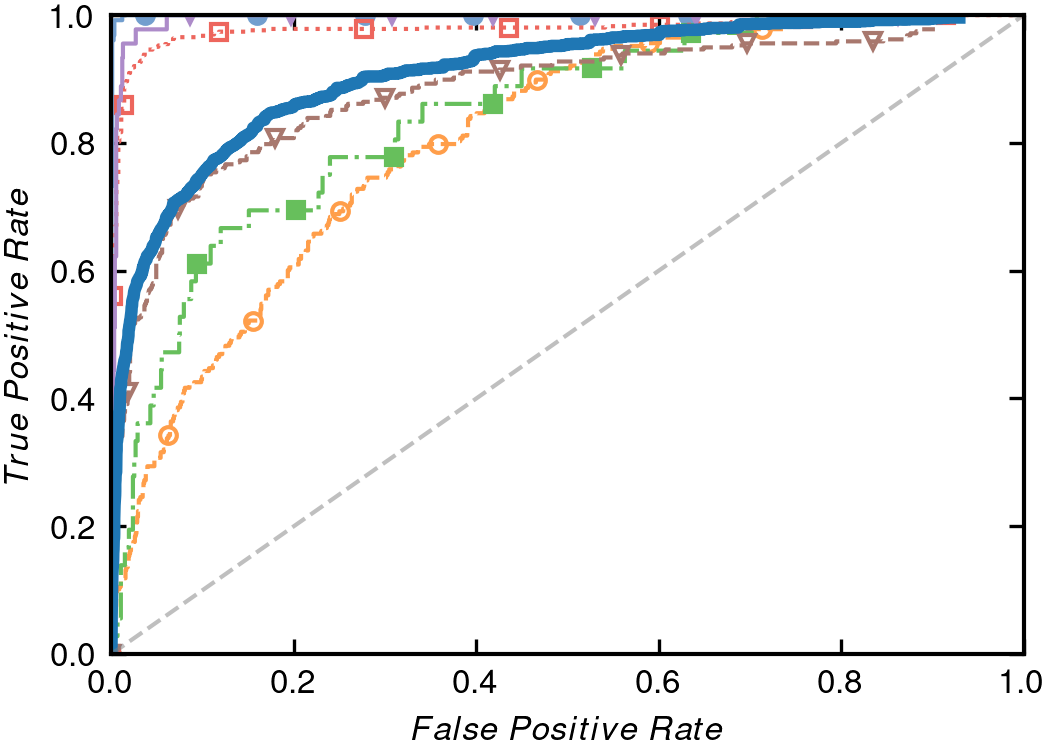}
\end{minipage}%
\hspace{0.02\textwidth}%
\begin{minipage}{0.45\textwidth}
  \includegraphics[width=\linewidth]{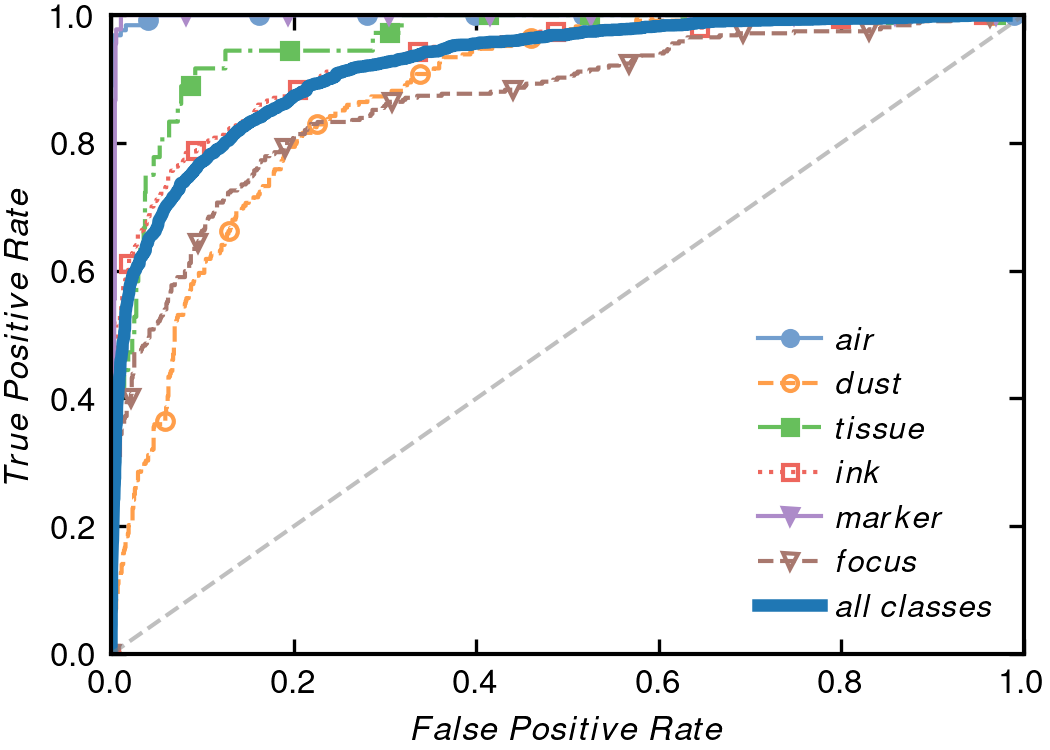}
\end{minipage}%
\caption{ROC curve for classification, evaluated on additional ACROBAT annotations unseen during training. (left) model trained on $\mathbf{ACR}$. (right) model trained on augmented $\mathbf{ACR'}$.}
\label{fig:classification-evalacr}
\end{figure}

\begin{figure}[!h]
\centering%
\includegraphics[width=0.6\textwidth]{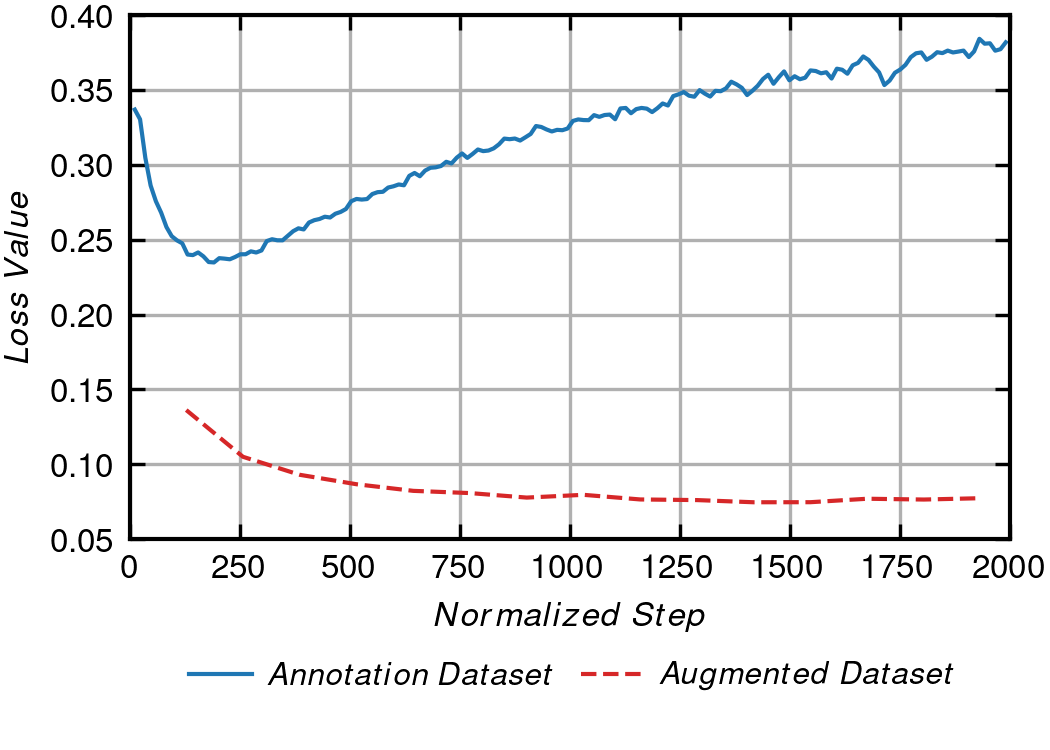}
\caption{Chart of the validation loss during training on $\mathbf{ACR}$ and $\mathbf{ACR'}$ datasets.}
\label{fig:loss-evalacr}
\end{figure}

The classification study is presented through Receiver Operating Characteristic (ROC) curves accompanied by their corresponding Areas Under the Curve (AUC) scores (Table~\ref{tab:acc_summary}), offering a comprehensive evaluation of the models' performance. Figure~\ref{fig:classification-evalacr} illustrates the promising initial validation results, with improvements evident when employing the augmented dataset, particularly in addressing previously weaker outcomes. Figure~\ref{fig:loss-evalacr} details the loss on the validation dataset, highlighting the mitigation of overfitting issues with the augmented dataset during the training process. Subsequent testing on additional ACROBAT annotations reveals improvements for tissue and dust artifacts, alongside a performance decrease for ink artifacts and a slight drop in focus artifacts. Evaluation on ANHIR annotations demonstrates enhancements for air, tissue, dust, and focus artifacts, tempered by slight degradation in marker and ink artifacts.

\begin{figure}[!h]
\centering%
\begin{minipage}{0.45\textwidth}
  \includegraphics[width=\linewidth]{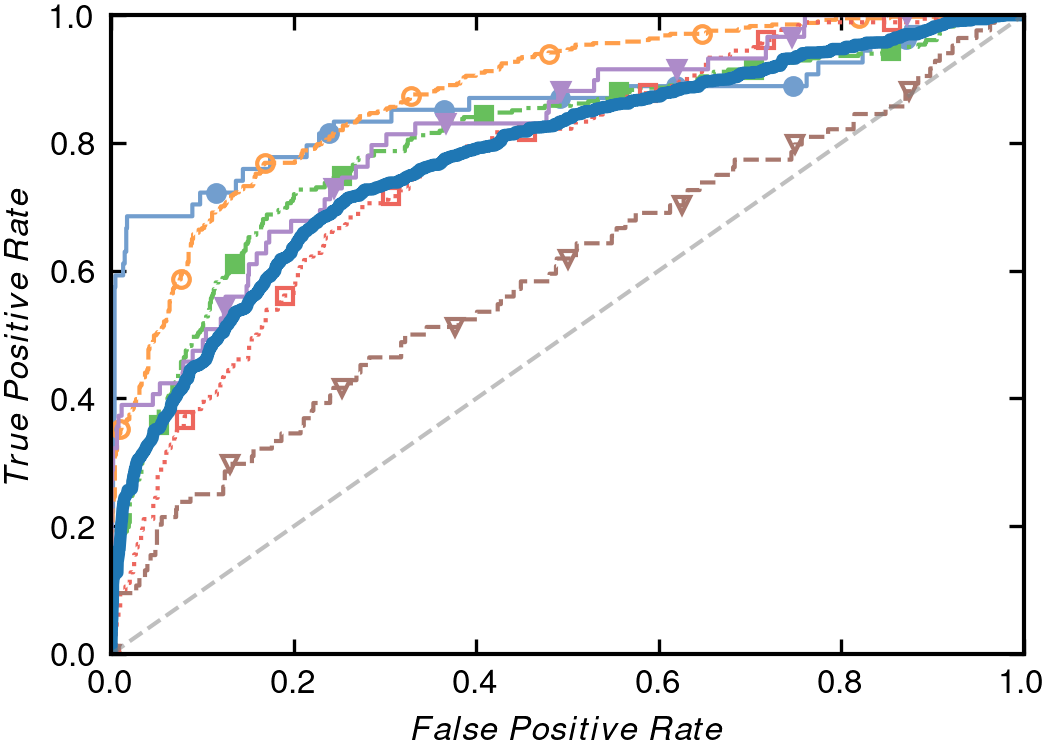}
\end{minipage}%
\hspace{0.02\textwidth}%
\begin{minipage}{0.45\textwidth}
  \includegraphics[width=\linewidth]{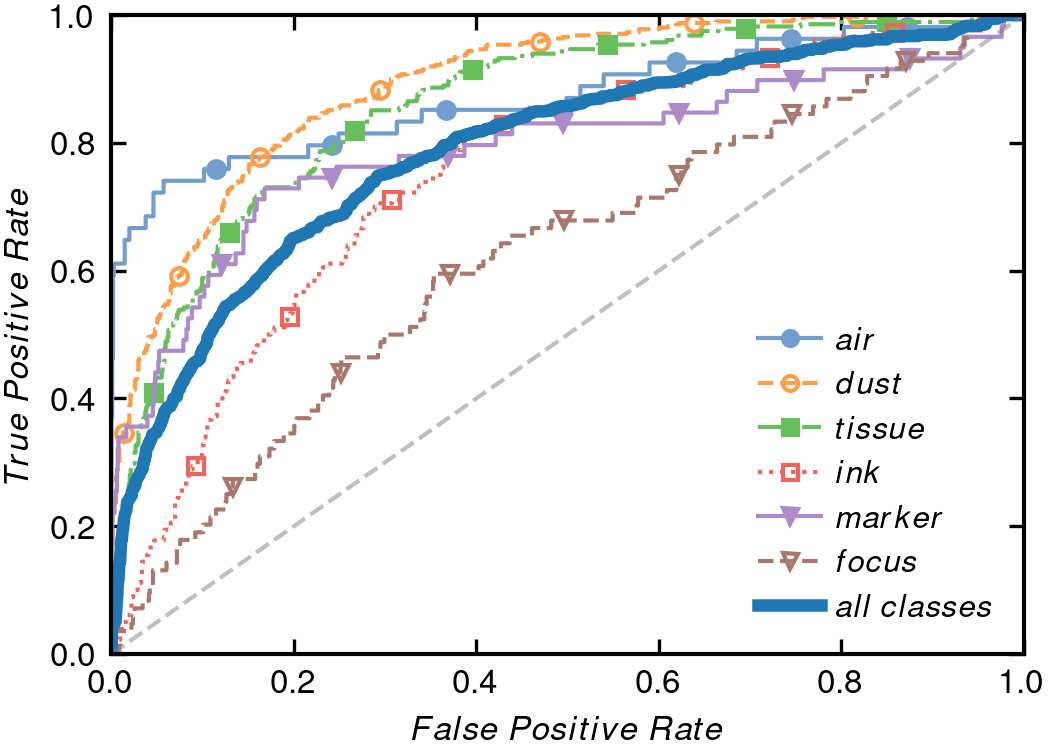}
\end{minipage}%
\caption{ROC curve for classification models, evaluated on ANHIR annotations unseen during training. (left) model trained on $\mathbf{ACR_{anh}}$. (right) model trained on augmented $\mathbf{ACR'_{anh}}$.}
\label{fig:classification-evalanh}
\end{figure}

\begin{figure}[!h]
\centering%
\begin{minipage}{0.45\textwidth}
  \includegraphics[width=\linewidth]{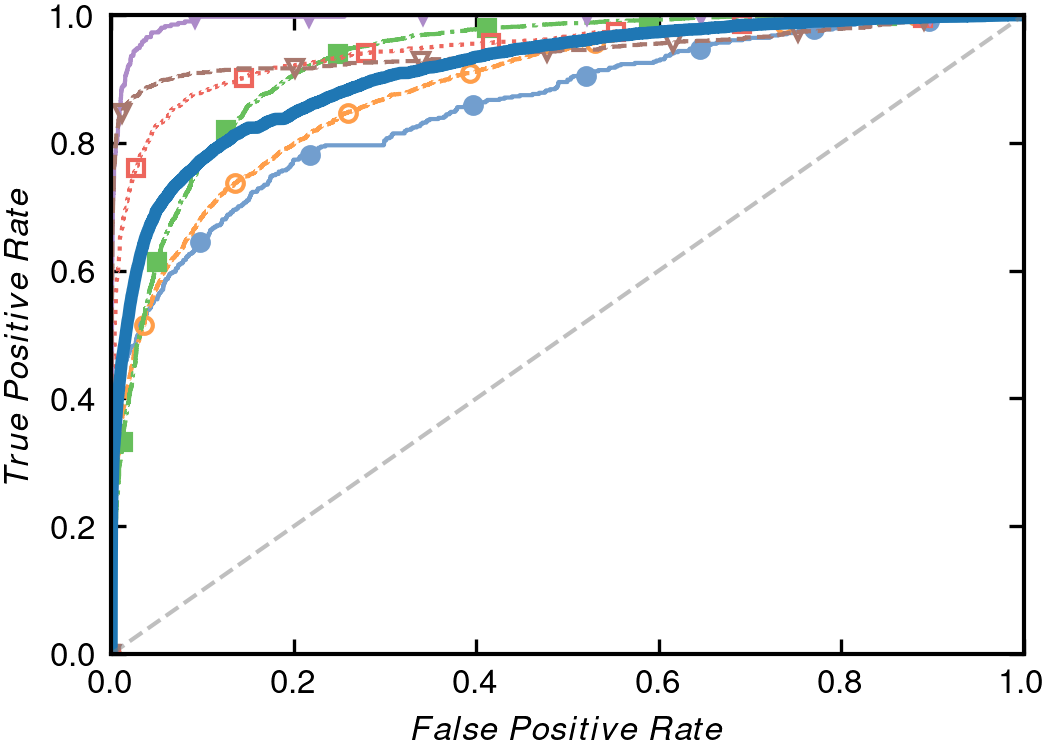}
\end{minipage}%
\hspace{0.02\textwidth}%
\begin{minipage}{0.45\textwidth}
  \includegraphics[width=\linewidth]{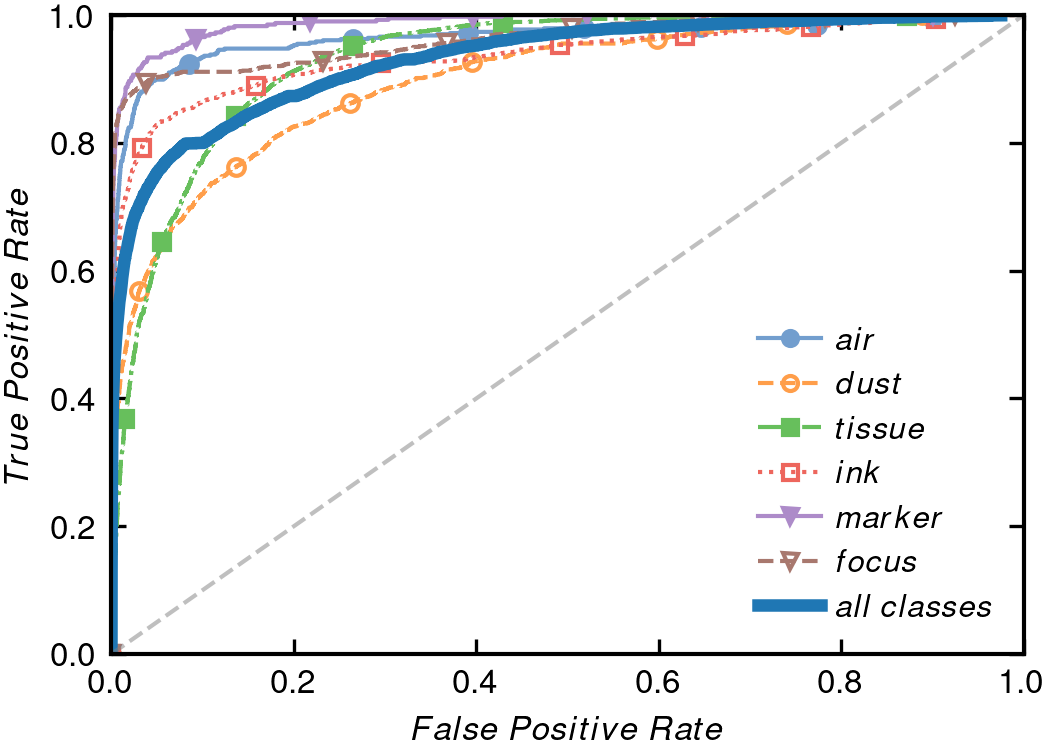}
\end{minipage}%
\caption{ROC curve for classification models, evaluated on Radboud University test annotations consisting of evenly sampled 70\% of all dataset annotations. (left) model trained on $\mathbf{RB}$. (right) model trained on augmented $\mathbf{RB'}$.}
\label{fig:classification-evalbp}
\end{figure}

In Figure~\ref{fig:classification-evalanh}, the models undergo evaluation on a diverse set of WSIs. The augmented dataset yields improvements for air and dust artifacts, with more significant enhancements for dust, focus, and tissue types. However, a slight degradation is observed for ink and marker types. Improvement in performance on this dataset is the lowest overall. Further analysis in Table~\ref{tab:acc_diff_summary} reveals that the model does not generalize well to a new dataset. The lack of statistical significance is confirmed by the statistical tests. Evaluation on the Radboud University dataset (Figure~\ref{fig:classification-evalbp}) demonstrates an overall improvement, notably for the initially weakest artifact---Air bubbles. Better results are also observed for tissue and focus, with marginal gains for dust and a slight regression for the ink class.

\begin{figure}[!h]
\centering%
\begin{minipage}{0.45\textwidth}
  \includegraphics[width=\linewidth]{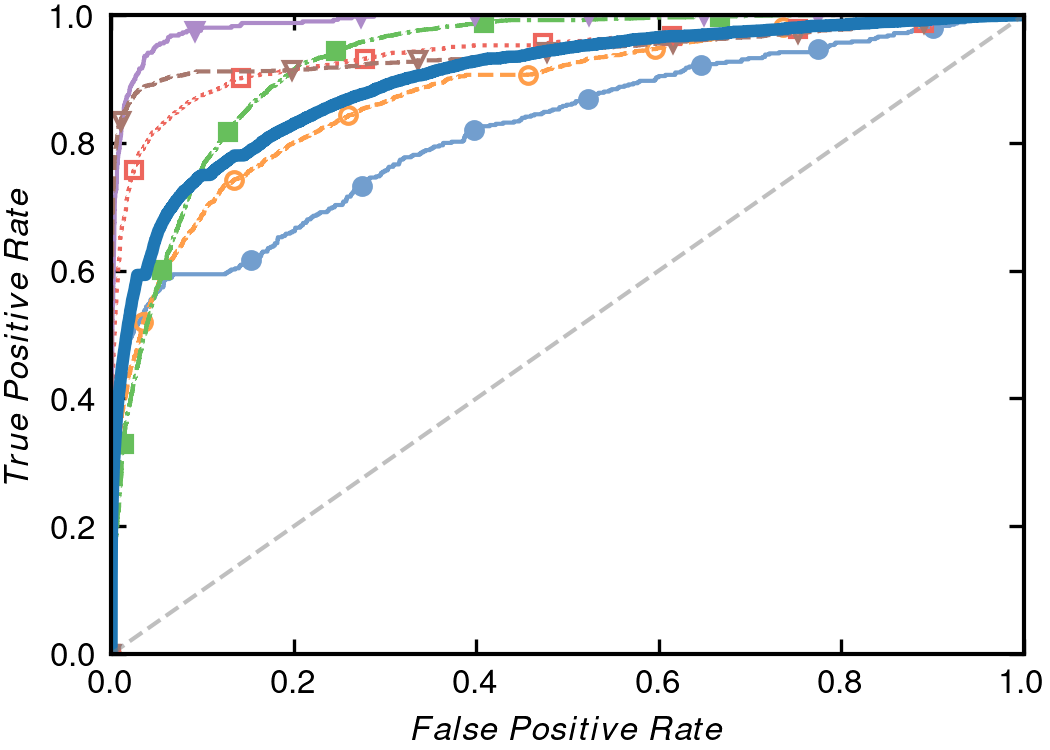}
\end{minipage}%
\hspace{0.02\textwidth}%
\begin{minipage}{0.45\textwidth}
  \includegraphics[width=\linewidth]{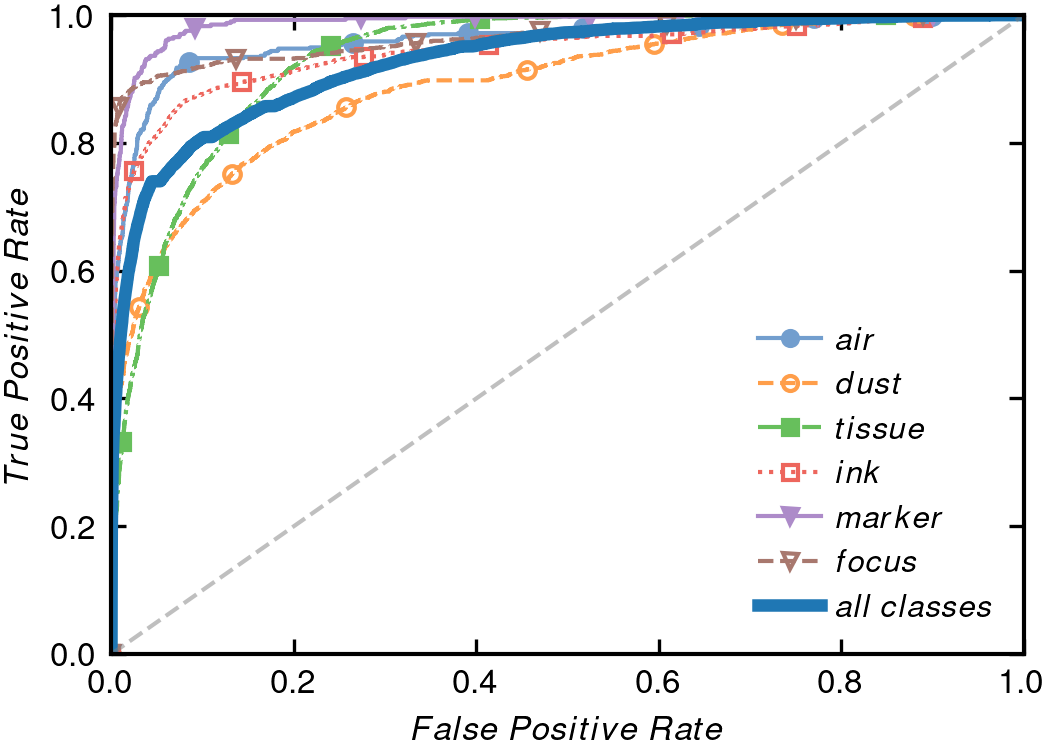}
\end{minipage}%
\caption{ROC curve for classification models, evaluated on Radboud University test annotations consisting of evenly sampled 70\% of all dataset annotations. Model training was limited to only the last fully connected layer. (left) model trained on $\mathbf{RB_s}$. (right) model trained on augmented $\mathbf{RB'_s}$.}
\label{fig:classification-shallow}
\end{figure}

Despite freezing all layers, except the last fully connected layer with 2048 input features and 6 output classes, some overfitting was observed, indicated by an initial increase in loss. However, the loss stabilized over time, and the final performance exhibited less degradation for previously challenging artifacts. The concluding experiment involving a mostly frozen model (Figure~\ref{fig:classification-shallow}) highlights the overfitting issue, with a noticeable performance drop for the dataset with only annotations as a training data. Notably, this regression is absent for the augmented dataset. Comparison between the two sets reveals that the model trained on the augmented dataset outperforms the model trained solely on annotations for all artifact types, indicating the effectiveness of the proposed augmentation approach in mitigating overfitting.

\begin{figure}[h]
\centering%
\begin{minipage}{0.45\textwidth}
  \includegraphics[width=\linewidth]{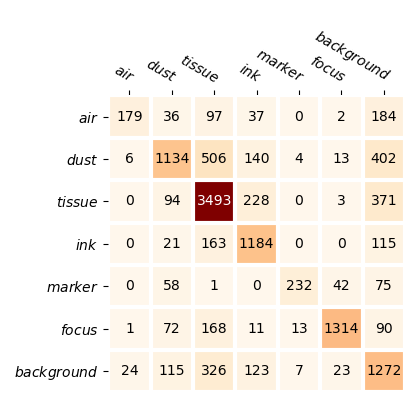}
\end{minipage}%
\hspace{0.02\textwidth}%
\begin{minipage}{0.45\textwidth}
  \includegraphics[width=\linewidth]{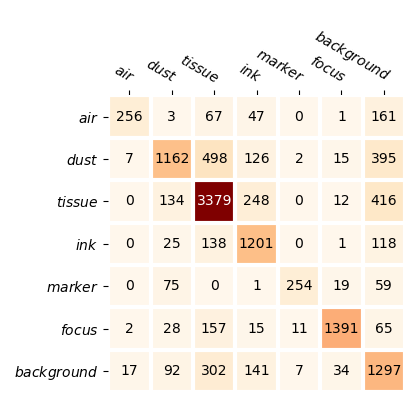}
\end{minipage}%
\caption{Confusion matrix of predictions after thresholding. (left) model trained on $\mathbf{RB}$. (right) model trained on augmented $\mathbf{RB'}$.}
\label{fig:classification-cm}
\end{figure}

In Figure~\ref{fig:classification-cm} we see the confusion matrix after thresholding. Background class was raised when no other class met the required threshold. The high values along the diagonal elements of the matrix indicate that our model was successful in correctly classifying instances across multiple classes. Nonetheless, we have observed patterns in misclassifications, with specific classes exhibiting higher rates of misclassifications, e.g., dust, focus, and background.

\begin{table}
    \centering
    \caption{Summary of the final performance of the models on each artifact type defined by the AUROC score.}
    \begin{tabular}{ccccccccc}
        \hline
         \textbf{Artifact Type} & $\mathbf{ACR}$ & $\mathbf{ACR'}$ & $\mathbf{ACR_{anh}}$ & $\mathbf{ACR'_{anh}}$ & $\mathbf{RB}$ & $\mathbf{RB'}$ & $\mathbf{RB_s}$ & $\mathbf{RB'_s}$ \\
         \hline
        \textbf{Air}         & 1.000 & 0.999 & 0.858 & 0.872 & 0.858 & 0.965 & 0.820 & 0.961 \\
        \textbf{Dust}        & 0.807 & 0.876 & 0.876 & 0.892 & 0.887 & 0.898 & 0.880 & 0.889 \\
        \textbf{Tissue}      & 0.843 & 0.957 & 0.803 & 0.859 & 0.928 & 0.932 & 0.929 & 0.934 \\
        \textbf{Ink}         & 0.978 & 0.930 & 0.770 & 0.753 & 0.947 & 0.936 & 0.940 & 0.943 \\
        \textbf{Marker}      & 0.995 & 0.998 & 0.820 & 0.791 & 0.994 & 0.988 & 0.989 & 0.989 \\
        \textbf{Focus}       & 0.880 & 0.867 & 0.597 & 0.621 & 0.945 & 0.963 & 0.944 & 0.966 \\
        \textbf{All classes} & 0.917 & 0.938 & 0.787 & 0.798 & 0.927 & 0.947 & 0.917 & 0.947 \\
        \hline
    \end{tabular}
    \label{tab:acc_summary}
\end{table}

\begin{table}
    \centering
    \caption{Summary of the improvements in AUROC made by our method in each dataset and for each artifact type. All differences are presented with an additional Wilcoxon signed-rank test performed on an accumulated list of patch predictions.}
    \begin{tabular}{ccccc}
        \hline
         \textbf{Artifact Type} & $\mathbf{ACR'-ACR}$ & $\mathbf{ACR'_{anh}-ACR_{anh}}$ & $\mathbf{RB'-RB}$ & $\mathbf{RB'_s-RB_s}$ \\
         \hline
        \textbf{Air}         & -0.001 &  0.014 &  0.107 &  0.141 \\
        \textbf{Dust}        &  0.069 &  0.016 &  0.011 &  0.009 \\
        \textbf{Tissue}      &  0.114 &  0.055 &  0.004 &  0.005 \\
        \textbf{Ink}         & -0.049 & -0.017 & -0.011 &  0.003 \\
        \textbf{Marker}      &  0.003 & -0.029 & -0.006 & -0.001 \\
        \textbf{Focus}       & -0.013 &  0.024 &  0.018 &  0.021 \\
        \textbf{All classes} &  0.021 &  0.011 &  0.020 &  0.030 \\
        \textbf{pvalue}      &  1E-03 &  3E-01 &  9E-36 &  4E-06 \\
        \hline
    \end{tabular}
    \label{tab:acc_diff_summary}
\end{table}

\section*{Discussion}

The presented data augmentation pipeline, blending annotated artifacts from a donor dataset to the destination, demonstrates its utility and positive impact on artifact detection. This method proves effective in generalizing a QC system to real-world histology data, even with limited training annotations, substantially increasing their quantity. Notably, the method excels in handling underrepresented or highly heterogeneous artifacts, evident in significant improvements for artifacts like air and dust. Improvements in handling focus distortions, which are relatively easier to generate correctly, were also observed. The method effectively addresses tissue artifacts, particularly in datasets with similar staining characteristics, though a slight degradation is noted for a complete mix of datasets. This comes from an improved heterogeneity of the training dataset, and could be further improved by increasing the initial sample of the annotated artifacts. 

The method's handling of ink artifacts presents a challenge, with potential future improvements in stain transfer methods and the incorporation of advancements from generative deep learning methods. Additionally, the focus artifact type is often mixed with tissue folding. This may be attributed to the frequent presence of both types of artifacts together, as the folds have the potential to cause difficulties with focus. The issue with the background classification could be related to the fact that not all the artifacts present were annotated, and there are patches labeled as containing none of the artifacts when in fact there are.

The study also indicates that the proposed method can mitigate overfitting in high-performing models, as evident in the analysis of validation loss graphs. Quality tissue segmentation significantly influences blur detection, with notable increases in accuracy when proper segmentation is achieved.

Future work could focus on the classification network, exploring potential shortcomings in handling small, irregular shapes or closely situated objects. The study suggests improvements in handling ink artifacts and investigating advanced filtering on the edges of inserted marker artifacts. Further research could explore recent advancements in generative deep learning methods for stain transfer and expand the artifact types, requiring more collaboration with professional pathologists and incorporating additional datasets for a more diverse artifact collection.

\section*{Methods}

\subsection*{Overview}

The proposed framework addresses the challenge of augmenting Whole Slide Images (WSIs) with artifacts through an optimized processing pipeline. To efficiently handle large-scale histopathology datasets, the framework adopts a streaming approach for augmentation, loading patches sequentially to overcome memory constraints. Images are saved in the \textit{.tiff} format, compatible with the standard pyramidal image structure of OpenSlide---an open standard in histopathology, ensuring compatibility with prevalent tools \cite{goode_openslide_2013, goode_vendor-neutral_2007}. Annotations from the artifact detection framework are stored in the \textit{.xml} format, chosen for its human-readable nature and compatibility with ASAP (Automated Slide Analysis Platform) software \cite{noauthor_asap_2023}.

To leverage high-resolution capabilities without downsampling, the framework strategically reads only the Region of Interest (ROI) instead of the entire image. Focusing solely on areas with artifacts maximizes resolution utilization, ensuring no loss of critical details during the detection process. The pipeline begins by referencing the image and corresponding annotations, with the image serving as the source for artifact extraction and annotations providing vital information. This iterative extraction process establishes the groundwork for subsequent processing steps, ensuring accuracy in artifact identification.

\begin{figure}[!h]
\centering%
\includegraphics[width=\linewidth]{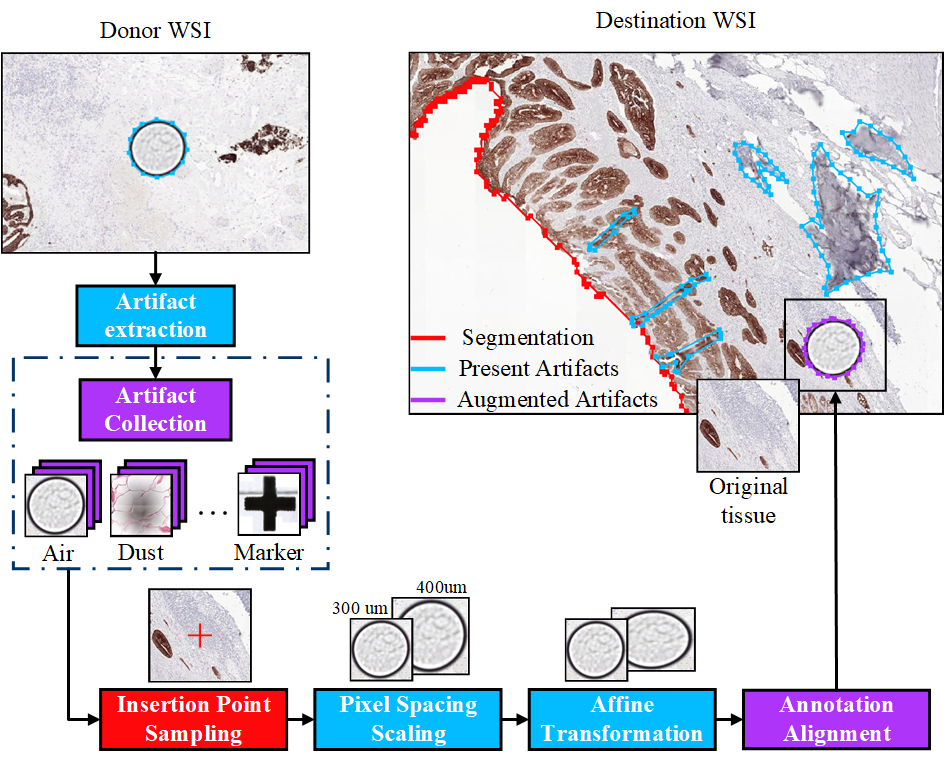}
\caption{The process of augmenting a WSI with an artifact coming from previously extracted artifact collection.}
\label{fig:library}
\end{figure}

\begin{figure}[!h]
\centering%
\includegraphics[width=\linewidth]{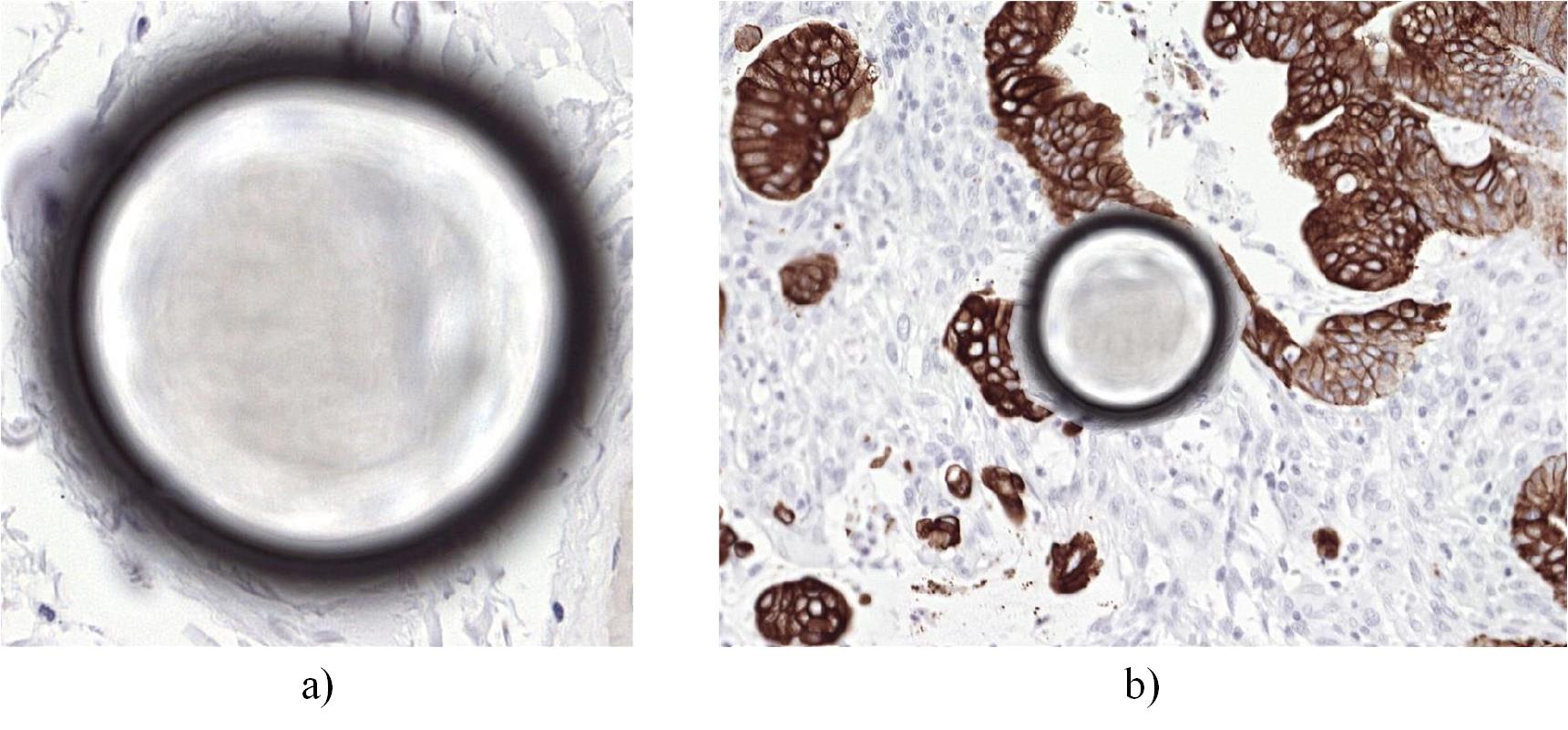}
\caption{Example of an air bubble being blended onto the destination WSI: a) original image, b) artifact blended with the destination image}
\label{fig:example-air}
\end{figure}

Following the extraction of artifacts from the Region of Interest (ROI), the pipeline stores each artifact in a dedicated collection, referred to as the Artifact Collection. This repository retains all detected artifacts, serving as a reference for subsequent stages. The augmentation process, illustrated in Figure~\ref{fig:library}, iteratively blends these artifacts onto the destination WSI. Throughout this augmentation process, the following sequential steps are executed:

\begin{itemize}
\item \textbf{Sampling of the Artifact Insertion Point:} The central point for artifact insertion is determined based on the specific artifact type, with detailed configurations provided in Table~\ref{tab:config}.
\item \textbf{Scaling the Artifact to Correct Pixel Spacing:} This critical step adjusts artifacts originating from different scanners or tissues to maintain real-life measurements. File metadata is utilized to preserve the physical measurements of each WSI's pixels.
\item \textbf{Affine Transformation:} To diversify the resulting data, artifacts undergo augmentation, incorporating randomly applied rotation and scaling. Affine transformation is selected for its avoidance of tearing or folding artifacts.
\item \textbf{Aligning the Annotation:} Following all applied operations on the artifact image, its annotation undergoes an identical transformation to ensure alignment with the augmented artifact.
\end{itemize}

\begin{table}[!ht]
    \centering
    \caption{Summary of the used configuration for each artifact type.}
    \begin{tabular}{ccc}
    \hline
        \textbf{Artifact Type} & \textbf{Max no. of inserted artifacts} & \textbf{Location of inserted artifacts} \\ \hline
        \textbf{Air} & 4 & whole WSI \\ 
        \textbf{Dust} & 7 & whole WSI \\ 
        \textbf{Tissue} & 4 & on top of the tissue (foreground) \\ 
        \textbf{Ink} & 4 & edge of the tissue \\ 
        \textbf{Marker} & 4 & outside the tissue (background) \\ 
        \textbf{Focus} & 2 & on top of the tissue (foreground) \\ \hline
    \end{tabular}
    \label{tab:config}
\end{table}

To address the necessity for stain-invariant segmentation for insertion point sampling, we used a custom deep learning model. For detailed architecture and training pipeline specifics, readers are directed to our previous paper \cite{jurgas_robust_2024}. The method was selected after a thorough consideration of available options, acknowledging the critical need for a stain-invariant model capable of handling images coming from different datasets with diverse magnifications. This stain-invariant and multiresolution segmentation increases the overall robustness of the framework across different datasets and staining methodologies.

\begin{equation}
I_\text{out} = Gauss(M_\text{artifact}) \times I_\text{artifact} + (1-Gauss(M_\text{artifact})) \times I_\text{WSI}
\label{eq:blend}
\end{equation}

Moreover, the pipeline can cut the augmented WSI into patches. Leveraging the segmentation module, this allows for even sampling of each artifact type, while simultaneously preserving a specified number of empty, background patches. This balanced training approach contributes to fewer false positive predictions. To execute the blending process, distinct strategies are employed for each artifact group:

\begin{itemize}
\item \textbf{Focus Distortions (Gaussian Blurring):} Gaussian blurring simulates focus distortions in the blending process.
\item \textbf{Markers and Air Bubbles:} These artifacts are blended by insertion onto the destination image, followed by bilateral filtering.
\item \textbf{Dust Artifacts:} Investigated with seamless cloning (gradient editing) and the previously described method for markers.
\item \textbf{Ink Transfer:} Reinhard Color Normalization \cite{reinhard_color_2001} is employed to maintain the original tissue structure during ink transfer. Alternative methods such as \cite{macenko_method_2009} were explored, but did not yield superior results and increased computational costs.
\end{itemize}

After each blending step, further processing is applied to the edges of the inserted artifact. A smooth transition, described by Equation~\ref{eq:blend}, is implemented. This equation outlines a gradual, linear transition between pixel values of the artifact and the destination image based on a Gaussian-smoothed artifact's annotation mask. Figure~\ref{fig:example-air} visually depicts artifacts both before and after the blending process, showcasing the efficacy of the proposed methodology.

\subsection*{Classification}
Image patches were classified to evaluate potential challenges, especially for small, irregular shapes or closely situated objects \cite{guo_msft-yolo_2022}. Patches with artifacts and backgrounds of size $224\times224$ px were cut from the images. 

A pretrained ResNet50 model was selected for this classification task. In most experiments, the whole model was trained. In the case of $\mathbf{RB_s}$ and $\mathbf{RB'_s}$ datasets, the last convolutional layer and one fully connected layer at the end of the network were unfrozen to mitigate overfitting. The model's feasibility was assessed by training it on dataset annotations and then on an augmented dataset. Two evaluations were conducted: one with additional unseen annotation patches and another with annotation patches from an entirely new dataset to assess generalizability. Two additional experiments utilized professionally annotated data, training the model on annotations only and on an augmented dataset that included data from all three datasets. The experiment was evaluated on an unseen subsample of the professional annotations. The last experiment was repeated with a reduction of trainable model layers to only the last fully connected layer to assess potential overfitting.

\subsection*{Ethical approval}

For this study we are using existing data coming from two open datasets (ACROBAT and ANHIR) which received ethical approval from their respective institutional review boards. For data coming from the Radboud University, all methods were carried out in accordance with relevant guidelines and regulations. Experimental protocols and patient inclusion were approved by the ethical review board (METC) of Radboudumc, Nijmegen, Netherlands. For all included patients, informed consent was obtained.

\section*{Conclusions}
In conclusion, our QC system demonstrates efficacy in accurate artifact detection and classification in histopathology images. Utilizing annotated datasets with advanced data augmentation techniques promises improved performance and reduced overfitting. While challenges in handling specific artifacts and adapting to various magnification levels and datasets exist, this study contributes to developing a reliable histopathology QC system for enhanced image analysis and accurate clinical diagnosis. We release the code freely at \cite{jurgas_jararturhistopathologyaugmentationresearch_2023}.

\bibliography{myreferences}

\section*{Acknowledgements}

This project has received funding from the Innovative Medicines Initiative 2 Joint Undertaking under grant agreement No 945358. This Joint Undertaking receives support from the European Union's Horizon 2020 research and innovation program and EFPIA, Belgium (www.imi.europe.eu). The research reflects only the author's view and the Joint Undertaking is not responsible for any use that may be made of the information it contains. Additionally, the research was supported in part by PLGrid Infrastructure. We gratefully acknowledge Poland’s high-performance computing infrastructure PLGrid (HPC Centers: ACK Cyfronet AGH) for providing computer facilities and support within computational grant no. PLG/2023/016239.

\section*{Author contributions statement}

A.J. and M.W. conceptualized the method, A.J. carried out the implementation, conducted the experiments and evaluation. Data acquisition was done by M.A (Marina D'Amato) and J.L. Results analysis was done by all authors. The article was prepared by A.J. and M.W. and reviewed by all the authors. J.L, M.A. (Manfredo Atzori), and H.M. supervised the project.

\end{document}